\documentclass[letterpaper]{article} 
\usepackage{aaai2027}  
\usepackage[hyphens]{url}  
\usepackage{graphicx} 
\usepackage{natbib}  
\usepackage{caption} 
\usepackage{algorithm}
\usepackage{algorithmic}

\usepackage{amsthm}
\usepackage{amsmath, bm}
\usepackage{amsfonts}
\usepackage{upgreek}
\usepackage{svg}
\usepackage{booktabs}
\usepackage{multirow}
\usepackage{subfig}
\usepackage{fontawesome5}
\usepackage[dvipsnames,table]{xcolor}
\usepackage{enumitem}
\usepackage{subcaption}

\usepackage{booktabs}
\usepackage{multirow}
\usepackage{graphicx}
\usepackage{amssymb}
\usepackage{threeparttable}

\usepackage{xcolor}
\usepackage{tabularx}
\usepackage{array}

\usepackage{makecell}

\newcommand{\first}[1]{\cellcolor{gray!15}{\textbf{#1}}}

\usepackage{newfloat}
\usepackage{listings}
\DeclareCaptionStyle{ruled}{labelfont=normalfont,labelsep=colon,strut=off} 
\floatstyle{ruled}
\newfloat{listing}{tb}{lst}{}
\floatname{listing}{Listing}

\usepackage{booktabs}

\title{Low-Cost Sensor Calibration for Indoor Air Quality Monitoring: A Dataset, Evaluation Scenarios, and a Lightweight Model}
\author {
    Jinyong Yun\textsuperscript{\rm 1},
    Seokho Ahn\textsuperscript{\rm 1},
    Hyungjin Kim\textsuperscript{\rm 1},
    Sungbok Shin\textsuperscript{\rm 2},
    Young-Duk Seo\textsuperscript{\rm 1}\thanks{Corresponding author.}
}
\affiliations {
    \textsuperscript{\rm 1}Department of Electrical and Computer Engineering, Inha University, Incheon 22212, South Korea\\
    \textsuperscript{\rm 2} Department of Computer Science and Engineering, Sogang University, Seoul 04107, South Korea\\
    12191632Y@inha.edu, sokho0514@inha.edu, flslzk@inha.edu,  sbshin90@sogang.ac.kr, mysid88@inha.ac.kr
}
\nocopyright

\begin{document}
\maketitle
\begin{abstract}
Low-cost sensors enable scalable indoor air quality monitoring but require calibration because of nonlinear distortions, noise, and temporal drift. The conventional strict pairwise calibration setting requires a co-located reference sensor at each deployment location and does not account for spatial and temporal heterogeneity. 
To address these limitations, we introduce a six-month dataset comprising multivariate indoor air-quality measurements from low-cost and reference sensors with contextual metadata collected at five locations.
Using this dataset, we define four evaluation scenarios. The reference-efficient and location-transfer scenarios evaluate spatial generalization, whereas the long-term drift and event-conditioned scenarios assess robustness to gradual and abrupt distribution shifts.
Based on these scenarios, we derive design requirements and propose a lightweight temporal model that combines input-window compression with residual temporal and feature fusion. Experiments show strong calibration performance across all four scenarios with low edge-inference cost.
\end{abstract}

\section{Introduction}
\label{sec:01_intro}

People spend about 90\% of their time indoors, making indoor environments a major source of daily pollutant exposure and an important factor in human health~\cite{jiang2011maqs, zhong2021complexity}. Accurate measurement of indoor air quality is therefore essential for protecting people. However, indoor environments change continuously due to short-term events such as ventilation, changes in occupancy, and human activities ~\cite{wu2021influence}. They also exhibit spatial heterogeneity, meaning that air quality can differ across locations even within the same indoor space~\cite{li2018spatiotemporal}. Because of this temporal and spatial heterogeneity, a small number of sensors may fail to capture both local changes and the overall air quality distribution. Therefore, precise monitoring of indoor environments requires the dense deployment of high-accuracy sensors throughout the monitored space.

Deploying a dense network of high-accuracy reference sensors (\textit{i.e}., high-cost sensors), however, is costly. Low-cost sensors can be deployed at lower cost, but their measurements are often affected by noise and drift, making them less accurate~\cite{koziel2025efficient}. Hence, instead of installing reference sensors at every measurement point, calibrating low-cost sensors with a lightweight deep learning model to achieve near-reference accuracy can provide an alternative that balances cost and performance~\cite{concas2021low, ahn24sendal, dalbah2026veli}.

Low-cost sensor calibration models are trained using data collected from reference sensors. 
Under this conventional setting, termed \textit{strict pairwise calibration}, at least one reference sensor is required at every location where a low-cost sensor is deployed~\cite{yu2020airnet,ahn2025real}. However, deploying reference sensors at all locations makes effective indoor air quality monitoring prohibitively expensive.

Accordingly, practical indoor air quality monitoring requires a calibration method that can calibrate low-cost sensors across multiple locations using only a few reference sensors. Such a method must be robust to both rapid environmental changes and spatial heterogeneity. Addressing these practical requirements involves three key challenges: constructing a dataset representative of indoor environments, defining realistic evaluation scenarios, and designing an efficient and robust calibration model. Specifically, we focus on the following:

\begin{itemize}[noitemsep, leftmargin=1em, topsep=0em]
\item \textbf{Long-term, multi-location data:} A sensor calibration dataset must capture spatial variations and temporal changes under limited reference-sensor availability.

\item \textbf{Realistic evaluation scenarios:} Calibration must operate robustly under real-world deployment conditions, including at locations without reference sensors and under short- and long-term distribution shifts.

\item \textbf{Efficient and robust calibration:} A lightweight calibration model must effectively generalize across locations and distribution shifts using limited reference data.
\end{itemize}


\begin{table*}[t]
\centering
\caption{Comparison of low-cost sensor calibration studies and datasets.
\textbf{T.} and \textbf{RH.} denote temperature and relative humidity.}
\label{tab:dataset_comparison}
\vspace{-0.5em}

\resizebox{\textwidth}{!}{%
\setlength{\tabcolsep}{8pt}
\begin{tabular}{@{}lcccccccc@{}}

\toprule

\multirow[c]{2}{*}[-0.8ex]{%
    \textbf{Dataset (Calibration Study)}} &
\multirow[c]{2}{*}[-0.8ex]{%
    \textbf{Modalities}} &
\multicolumn{3}{c}{\textbf{Dataset Properties}} &
\multicolumn{4}{c}{\textbf{Deployment Scenarios}} \\

\cmidrule(l{0.5mm}r{0.5mm}){3-5}
\cmidrule(l{0.5mm}r{0.5mm}){6-9}

& &
\shortstack[c]{Indoor.} &
\shortstack[c]{MultiLoc.} &
\shortstack[c]{Event.} &
\shortstack[c]{S1} &
\shortstack[c]{S2} &
\shortstack[c]{S3} &
\shortstack[c]{S4} \\

\cmidrule(l{0.5mm}r{0.5mm}){1-2}
\cmidrule(l{0.5mm}r{0.5mm}){3-5}
\cmidrule(l{0.5mm}r{0.5mm}){6-9}

UCI Air Quality \citep{aula2022evaluation} &
CO, NO$_2$, T., RH. &
\(\times\) &
\(\times\) &
\(\times\) &
\(\times\) &
\(\times\) &
\(\checkmark\) &
\(\times\) \\

Author-collected \citep{allka2023temporal} &
O$_3$, T. &
\(\times\) &
\(\checkmark\) &
\(\times\) &
\(\times\) &
\(\times\) &
\(\times\) &
\(\times\) \\

Author-collected \citep{ahn24sendal} &
PM$_{10}$ &
\(\checkmark\) &
\(\checkmark\) &
\(\times\) &
\(\times\) &
\(\times\) &
\(\times\) &
\(\times\) \\

Author-collected \citep{bachechi2024hypeair} &
NO, NO$_2$, O$_3$ &
\(\times\) &
\(\checkmark\) &
\(\times\) &
\(\times\) &
\(\times\) &
\(\checkmark\) &
\(\times\) \\

SensEURCity \citep{ahn2025real} &
PM$_1$, PM$_{2.5}$, PM$_{10}$ &
\(\times\) &
\(\checkmark\) &
\(\times\) &
\(\times\) &
\(\times\) &
\(\times\) &
\(\times\) \\

AQ-SDR \citep{dalbah2026veli} &
PM$_{2.5}$ &
\(\times\) &
\(\checkmark\) &
\(\times\) &
\(\checkmark\) &
\(\times\) &
\(\checkmark\) &
\(\times\) \\

\textbf{Indoor Sensor Calibration (\textsc{Ours})} &
\textbf{CO$_2$, CO, T., RH., PM$_1$} &
\first{\(\checkmark\)} &
\first{\(\checkmark\)} &
\first{\(\checkmark\)} &
\first{\(\checkmark\)} &
\first{\(\checkmark\)} &
\first{\(\checkmark\)} &
\first{\(\checkmark\)} \\

\bottomrule

\end{tabular}%
}

\vspace{-0.4em}

\begin{center}
\fontsize{6.8pt}{7pt}\selectfont
\textbf{Dataset properties}. Indoor. (indoor setting); MultiLoc.(multi-location data); Event. (event metadata).
\textbf{Symbols}. Supported (\(\checkmark\)); not supported (\(\times\)).
\end{center}

\vspace{-1em}
\end{table*}

To address these challenges, we propose a benchmark for indoor sensor calibration. First, we construct a six-month, multi-location indoor air quality dataset comprising synchronized measurements from low-cost and reference sensors deployed at five locations in a university laboratory (Section~\ref{sec:dataset}). 
The dataset contains measurements of five environmental variables (\textit{i.e}., CO$_2$, CO, temperature, humidity, and PM$_1$), along with metadata on door and window states, enabling the analysis of both gradual distribution shifts and rapid environmental changes. 
Based on this dataset, we define four evaluation scenarios that reflect limited reference-data availability, unseen-location generalization, and short- and long-term distribution shifts (Section~\ref{sec:calibration_scenarios}). Guided by these scenarios, we derive five design requirements and propose a calibration model designed to satisfy these requirements (Section~\ref{sec:04-method}).

Our main contributions are summarized as follows:
\begin{itemize}[noitemsep, leftmargin=1em, topsep=0em]
\item \textbf{Dataset}: We construct a six-month, five-location indoor sensor calibration dataset with synchronized low-cost and reference sensor measurements and event metadata.
\item \textbf{Scenarios}: We design four practical scenarios for effective indoor sensor calibration, in which many low-cost sensors across multiple locations are calibrated using limited reference data under spatial and temporal shifts.
\item \textbf{Model}: We propose a lightweight temporal calibration model that robustly addresses scenario-driven design requirements with low edge-inference cost.
\end{itemize}

\section{Related Work}\label{sec:02-relatedwork}
We review prior work on low-cost sensor calibration and the datasets used in previous calibration studies.

\subsection{Sensor Calibration Datasets}
Table~\ref{tab:dataset_comparison} compares representative calibration datasets. 
UCI Air Quality~\cite{air_quality_387, aula2022evaluation} contains long-term measurements from a single outdoor site.
The datasets used by \citet{allka2023temporal} and HypeAIR~\cite{bachechi2024hypeair} extend this setting to multiple outdoor locations or longer deployment periods, while SensEURCity~\cite{van2023senseurcity} and AQ-SDR~\cite{dalbah2026veli} provide large-scale outdoor measurements across multiple regions.
Among indoor datasets, SenDaL~\cite{ahn24sendal} includes measurements from three environments but is limited to deployment periods of 10 days.
None of these datasets combines synchronized measurements across multiple indoor locations with long-term coverage and event metadata.

\noindent \textbf{Our contribution.}
Our six-month, five-location indoor dataset combines paired low-cost and
reference sensor measurements with door, window, and occupancy metadata,
supporting all four indoor sensor calibration scenarios.

\subsection{Sensor Calibration Models}
Time-series models can be adapted to sensor calibration by
learning temporal patterns in low-cost measurements. Representative
backbones include linear, Multilayer perceptron (MLP)-based models,
and attention-based architectures~\cite{zeng2023dlinear,
chen2026xlinear,liu2024itransformer}.

Existing calibration methods address different deployment challenges. ~\citet{allka2023temporal} and HypeAIR~\cite{bachechi2024hypeair} focus on temporal patterns and long-term outdoor calibration. SenDaL~\cite{ahn24sendal} and TESLA~\cite{ahn2025real} target efficient edge deployment, while Veli~\cite{dalbah2026veli} performs unsupervised calibration without paired reference measurements. However, existing methods address these challenges separately rather than
jointly considering limited reference supervision, unseen-location
generalization, short- and long-term shifts, and efficiency.

\noindent \textbf{Our contribution.}
Our model combines temporal compression and residual fusion for robust calibration with limited references, unseen locations, and distribution shifts, while retaining low edge inference cost.

\begin{figure*}
    \centering
    \includegraphics[width=0.98\textwidth]{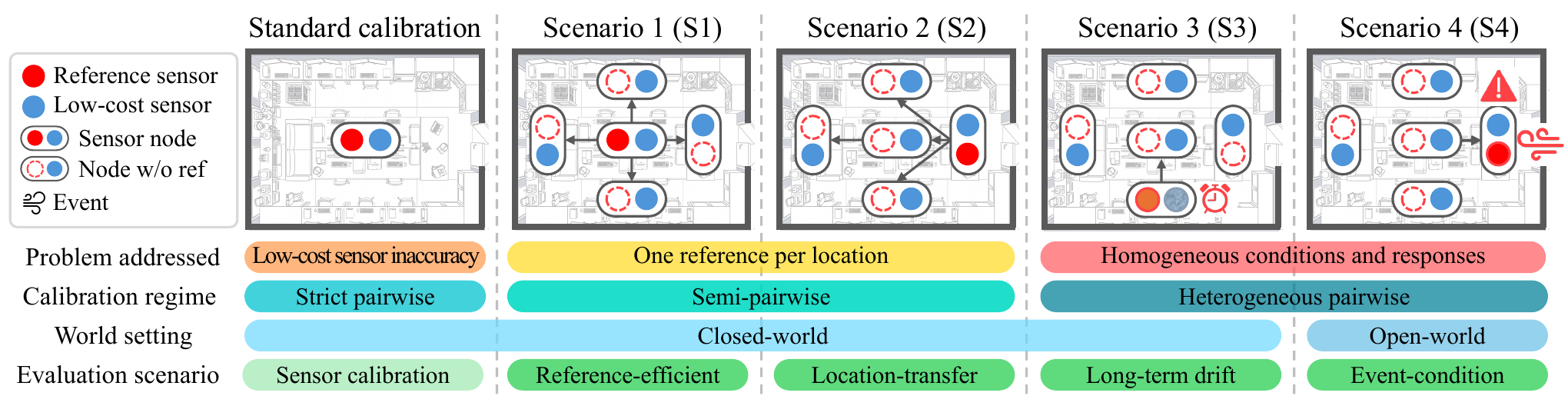}
    \vspace{-1em}
    \caption{Comparison of standard sensor calibration and four deployment-oriented scenarios in terms of relaxed assumptions, calibration regimes, and world settings:  \textbf{Scenarios 1-4:} reference-efficient, location-transfer, long-term drift, and event-condition.}
    \label{fig:Scenario}
    \vspace{-0.3em}
\end{figure*}

\section{Indoor Sensor Calibration Scenarios}
\label{sec:calibration_scenarios}

Figure~\ref{fig:Scenario} presents four evaluation scenarios that  relax \emph{strict pairwise calibration}, where each low-cost sensor is trained with synchronized, co-located reference sensor measurements. This conventional setting assumes: 

\begin{itemize}[noitemsep, leftmargin=1em, topsep=0em]
\item \textbf{One reference per location:}
Conventional sensor calibration requires a co-located reference sensor at each deployment location for model training. 
In indoor environments, 1) reference sensors are often placed only in selected zones, and 2) low-cost sensors may later be deployed at new zones not represented during training.

\item \textbf{Temporally stable calibration relationships:}
The relationship between low-cost and reference measurements is assumed to remain stable over time, overlooking 1) gradual sensor drift and 2) abrupt event-driven changes.
\end{itemize}

The four scenarios (S1-S4) directly implement the two cases under each assumption.
First, relaxing reference coverage (S1 and S2) is evaluated in the \emph{semi-pairwise} stage in Figure~\ref{fig:Scenario}, distinguished by whether the target location is observed during training:

\begin{itemize}[noitemsep, leftmargin=1.6em, topsep=0em]    
    \item [\textbf{S1}] \textbf{Reference-efficient.}
    Reference measurements are available only at reference-equipped (anchor) locations. Low-cost measurements from all locations are observed \textit{during} training.
    The model has the limited reference information across the observed network rather than generalizing to an unseen location.

    \item [\textbf{S2}] \textbf{Location-transfer.}
    The target location is excluded from training.
    The model learns from paired measurements at anchor locations and calibrates the unseen target using its low-cost measurements. Target references are used only for evaluation.
    Note that this scenario requires stronger spatial generalization than S1.
\end{itemize}

Next, relaxing temporal stability (S3 and S4) is evaluated in the \emph{heterogeneous-pairwise} stage, where each low-cost-reference sensor pair has a distinct calibration relationship.
Its two scenarios distinguish gradual internal drift from abrupt, externally induced shifts.

\begin{itemize}[noitemsep, leftmargin=1.6em, topsep=0em]
    \item [\textbf{S3}]\textbf{Long-term Drift.}
    The model is trained on an initial period and tested later, after sensor-specific responses have gradually changed.
    Because the shift occurs within the low-cost sensor network without abrupt external events, S3 is a \emph{closed-world} setting.
    
    \item [\textbf{S4}] \textbf{Event-condition.}
    External events, such as changes in door or window states, cause abrupt distribution shifts with location-dependent magnitudes and delays.
    The model must handle both temporal changes and location-specific responses.
    Because these shifts are externally induced, S4 is an \emph{open-world} setting.
\end{itemize}

\section{Indoor Sensor Calibration Dataset}
\label{sec:dataset}

To support the scenarios introduced in Section~\ref{sec:calibration_scenarios}, we construct a six-month calibration dataset with synchronized measurements from multiple locations, paired low-cost and reference sensor measurements, and event metadata.




\subsection{Data Collection}

The data is collected in a university laboratory from January 6 to July 6, 2026. 
We deploy sensor nodes at five locations within the same indoor space: East, West, South, North, and Center. 
All sensor nodes are installed at the same height of 0.75\,m above the floor to maintain consistent measurement conditions across locations.
The South node is located near a window, whereas the North node is located near an entrance, where they are more directly affected by ventilation and outdoor-air inflow. 
Together, these placements capture spatial variation within the room.

Each node records five environmental variables at 1-second intervals: CO$_2$, CO, PM$_1$, temperature (T.), and relative humidity (RH.).
For each variable, the low-cost and reference sensor are paired as listed in Table~\ref{tb:location_channel_statistics}.
We categorize the channels as low-cost and reference (high-cost) channels based on their
market prices at the time of data collection. 

After preprocessing, the dataset contains 261,597 1-minute location-timestamp observations.
Because the collection periods differ across locations, individual tasks contain between 223,460 and 260,483 paired observations, yielding 6,102,195 paired observations across all 25 tasks.

\begin{table}[t]
\centering
\vspace{0.25em}
\caption{Mean channel measurements across five locations after sanity
filtering and 1-minute resampling.}
\label{tb:location_channel_statistics}
\vspace{-0.5em}
\scriptsize
\setlength{\tabcolsep}{3pt}
\renewcommand{\arraystretch}{0.72}

\resizebox{\columnwidth}{!}{%
\begin{tabular}{clccccc@{}}
\toprule
\textbf{Measured}
&
\textbf{Channel}
&
\textbf{East}
&
\textbf{West}
&
\textbf{South}
&
\textbf{North}
&
\textbf{Center}
\\
\midrule

PM$_1$
&
\textcolor{blue!70!black}{\textbf{L}} PPD42
&
22.2 & 21.9 & 23.6 & 25.7 & 23.4
\\
($\mu$g/m$^3$)&
\textcolor{red!70!black}{\textbf{R}} PMS7003
&
20.2 & 18.9 & 21.4 & 21.8 & 19.6
\\

\midrule

CO
&
\textcolor{blue!70!black}{\textbf{L}} MQ135
&
6.3 & 7.1 & 6.2 & 8.5 & 6.8
\\
(ppm)&
\textcolor{red!70!black}{\textbf{R}} MQ9
&
4.5 & 4.2 & 4.4 & 4.8 & 4.1
\\

\midrule
CO$_2$
&
\textcolor{blue!70!black}{\textbf{L}} MQ135
&
552.8 & 540.0 & 548.0 & 584.0 & 556.2
\\
(ppm)&
\textcolor{red!70!black}{\textbf{R}} MG811
&
532.4 & 512.0 & 538.0 & 548.0 & 522.6
\\

\midrule
T.
&
\textcolor{blue!70!black}{\textbf{L}} DHT11
&
24.6 & 25.8 & 26.0 & 25.7 & 25.5
\\
($^\circ$C)&
\textcolor{red!70!black}{\textbf{R}} DHT22
&
24.1 & 24.2 & 24.4 & 24.1 & 23.2
\\

\midrule
RH.
&
\textcolor{blue!70!black}{\textbf{L}} DHT11
&
32.7 & 34.1 & 35.5 & 33.8 & 33.1
\\
(\%)&
\textcolor{red!70!black}{\textbf{R}} DHT22
&
33.8 & 34.0 & 34.5 & 33.1 & 32.9
\\

\bottomrule
\end{tabular}%
}

\vspace{0.15em}
\parbox{\columnwidth}{%
\tiny
\textcolor{blue!70!black}{\textbf{L}}: low-cost sensor;
\textcolor{red!70!black}{\textbf{R}}: reference sensor;
\textbf{T.}: Temperature;
\textbf{RH.}: Relative Humidity
}
\vspace{-1.5em}
\end{table}
\begin{figure*}[t]
    \centering
    \includegraphics[width=0.99\textwidth]{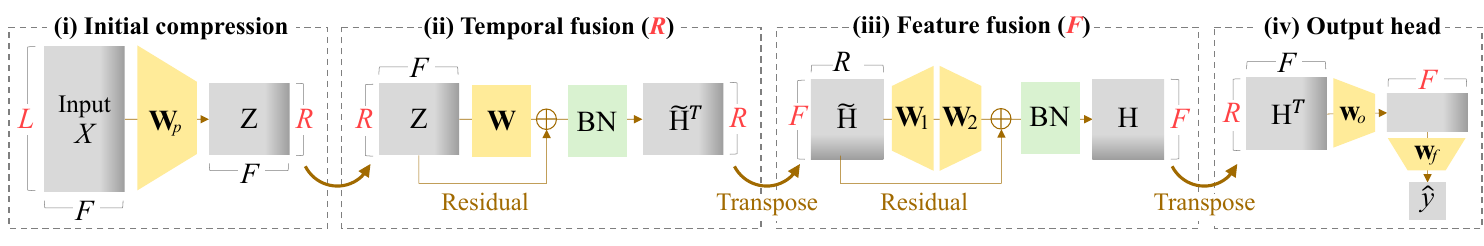}
    \vspace{-1em}
    \caption{Architecture of proposed model with initial temporal compression, temporal and feature fusion, and output head.}
    \label{fig:sensor_node_location}
\end{figure*}
\subsection{Metadata}
The dataset provides UTC and local timestamps, entrance-door state, the number of open windows (maximum 4), occupant count (maximum 11) during the preceding 30 minutes, and business-hour status. 
Door state and the number of open windows are recorded at 30-minute intervals. 
The metadata covers 223,460 1-minute timestamps.
At this 30-minute annotation resolution, we observe 4,118 door-state changes and 6,621 changes in the number of open windows. 
These values count state changes between consecutive valid annotations. 
The benchmark uses the door and window variables to identify event-conditioned evaluation periods, while the remaining metadata are provided for additional analysis.

\subsection{Preprocessing}

We process the data in the following manner. 
To begin with, for each location, timestamps were converted to UTC and sorted chronologically. 
Measurements sharing the same timestamp were aggregated by averaging each numerical channel. 
Nonnumeric and infinite values were treated as missing. 
Negative measurements were also treated as missing, except for temperature, for which negative values may be physically valid. 
Extreme values in the low-cost channels were removed using a relaxed filter based on the 1st and 99th percentiles. 
The 1-second measurements were then resampled into 1-minute averages.

Before imputation, valid-value ratios ranged from 82.9\% to 89.2\% across locations and channels. 
As summarized in Table~\ref{tb:location_channel_statistics}, channel means differed
across locations, including the window-side and entrance-side nodes.
These differences indicate that the calibration relationship depends on sensor placement and environmental conditions and cannot be represented by a single location-invariant offset.
Detailed missingness statistics and imputation procedures are provided in the Supplementary Material.

\subsection{Support for Deployment Scenarios}
The dataset was designed to support S1-S4 along two deployment
dimensions: spatial reference coverage and temporal stability. First, because synchronized low-cost and reference measurements are available at all five locations, we can simulate sparse-reference deployments by using reference measurements from only selected locations while retaining the same environmental data. Because the measurements share aligned timestamps and a common acquisition protocol, the dataset supports controlled comparisons between S1, which shares limited reference supervision across observed locations, and S2, which transfers
calibration relationships to a held-out location. In addition, the location-specific channel statistics in Table~\ref{tb:location_channel_statistics}, particularly at the South node near the window and the North node near the entrance, show that these scenarios capture genuine spatial heterogeneity rather than arbitrary partitions of otherwise homogeneous data.

Second, the continuous six-month time series preserves the temporal history of each sensor-location pair. Because each pair was continuously monitored for
six months, we can track how the calibration performance of the
same deployed sensor changes over time. This enables S3 to evaluate
gradual sensor drift. For S4, the distinct placements of the South and North nodes provide clear exposure to window ventilation and entrance-door air inflow, respectively. Together with repeated records of door- and window-state changes, these placements provide multiple event episodes for evaluating abrupt, delayed, and location-dependent sensor responses.

\begin{table*}[t]
\centering
\caption{Performance comparison on the representative outdoor
calibration dataset SensEURCity and our indoor dataset. The best result
for each metric is marked in \first{bold}, and the second-best result
is \underline{underlined}. Lower values indicate better performance.}
\label{tb:outdoor_indoor_results}
\vspace{-0.5em}
\footnotesize
\setlength{\tabcolsep}{8pt}

\resizebox{\textwidth}{!}{%
\begin{tabular}{@{}lcccccccccc@{}}
\toprule

\multirow{5}{*}{\textbf{Model}}
&
\multicolumn{3}{c}{\textbf{Outdoor data}}
&
\multicolumn{7}{c}{\textbf{Indoor data (Ours)}}
\\

\cmidrule(lr){2-4}
\cmidrule(lr){5-11}

&
\multicolumn{3}{c}{
    \textbf{RMSE} $\downarrow$
}
&
\multicolumn{5}{c}{
    \textbf{RMSE} $\downarrow$
}
&
\multirow{3.5}{*}{%
    \makecell[c]{%
        \textbf{Inference} $\downarrow$\\[-1pt]
        {\scriptsize \(\left(\mathrm{ms}\right)\)}
    }%
}
&
\multirow{3.5}{*}{%
    \makecell[c]{%
         \textbf{Memory} $\downarrow$\\[-1pt]
         {\scriptsize \(\left(\mathrm{KB}\right)\)}
    }%
}
\\

\cmidrule(lr){2-4}
\cmidrule(lr){5-9}


&
\makecell[c]{%
    \(\mathrm{PM}_{10}\)\\[-1pt]
    {\scriptsize
    \(\left(\mu\mathrm{g}/\mathrm{m}^{3}\right)\)}
}%
&
\makecell[c]{%
    \(\mathrm{PM}_{2.5}\)\\[-1pt]
    {\scriptsize
    \(\left(\mu\mathrm{g}/\mathrm{m}^{3}\right)\)}
}
&
\makecell[c]{%
    \(\mathrm{PM}_{1}\)\\[-1pt]
    {\scriptsize
    \(\left(\mu\mathrm{g}/\mathrm{m}^{3}\right)\)}
}
&
\makecell[c]{%
    \(\mathrm{CO}_{2}\)\\[-1pt]
    {\scriptsize
    \(\left(\mathrm{ppm}\right)\)}
}
&
\makecell[c]{%
    \(\mathrm{CO}\)\\[-1pt]
    {\scriptsize
    \(\left(\mathrm{ppm}\right)\)}
}
&
\makecell[c]{%
    \(\mathrm{PM}_{1}\)\\[-1pt]
    {\scriptsize
    \(\left(\mu\mathrm{g}/\mathrm{m}^{3}\right)\)}
}
&
\makecell[c]{%
    T.\\[-1pt]
    {\scriptsize \(\left({}^{\circ}\mathrm{C}\right)\)}
}
&
\makecell[c]{%
    RH.\\[-1pt]
    {\scriptsize \(\left(\%\right)\)}
}
&
&
\\

\midrule

iTransformer
& 15.92
& 7.36
& 4.07
& 9.80
& \underline{4.92}
& 8.54
& 2.43
& 5.24
& 2.32
& \underline{41.20}
\\

TimeXer
& 15.12
& \underline{6.31}
& 3.31
& \underline{9.05}
& 7.73
& 9.77
& 3.75
& 5.63
& 2.99
& 186.60
\\

DLinear
& 17.81
& 8.41
& 4.49
& 13.70
& 5.93
& 9.21
& 3.14
& 5.44
& \underline{1.29}
& \first{28.00}
\\

XLinear
& 19.05
& 11.64
& 8.48
& 14.71
& 5.92
& 8.71
& 3.23
& 6.20
& 2.12
& 152.00
\\

TimeMixer
& 16.75
& 8.72
& 4.69
& 19.67
& 6.84
& 10.07
& 3.27
& 8.46
& 5.84
& 384.00
\\

SenDaL
& 15.76
& 6.87
& 3.14
& 11.25
& 6.25
& 9.28
& 2.81
& 6.72
& 30.14
& 9614.00
\\

TESLA
& \underline{14.97}
& 6.55
& \underline{3.06}
& 9.06
& 5.36
& \underline{8.50}
& \first{1.96}
& \underline{5.07}
& 2.06
& 121.00
\\

\textbf{\textsc{Ours}}
& \first{13.34}
& \first{5.19}
& \first{2.50}
& \first{8.65}
& \first{4.80}
& \first{8.34}
& \underline{2.11}
& \first{4.89}
& \first{1.27}
& 42.60
\\

\bottomrule
\end{tabular}%
}

\vspace{-0.5em}
\end{table*}
\section{Model for Indoor Sensor Calibration}
\label{sec:04-method}
Based on our benchmark, this section defines key model design requirements for robust calibration across all scenarios (Section~\ref{subsec:design-req}) and presents a model designed to satisfy them (Section~\ref{subsec:method}).



\subsection{Design Requirements}
\label{subsec:design-req}
We define the following design requirements for robust indoor sensor
calibration across the four scenarios.

\begin{itemize}[noitemsep, leftmargin=1.8em, topsep=0em]    
    \item [\textbf{R1}] \textbf{Reference-efficient learning (S1).}
    The model should share limited reference supervision across multiple
    low-cost sensor locations.

    \item [\textbf{R2}] \textbf{Location-agnostic generalization (S2).}
    The model should learn transferable calibration patterns applicable
    to locations unseen during training.

    \item [\textbf{R3}] \textbf{Drift-robust representation (S3).}
    The model should remain robust to gradual distribution shifts in
    sensor baselines and responses over time.

    \item [\textbf{R4}] \textbf{Event-robust modeling (S4).}
    The model should robustly handle abrupt, delayed, and transient sensor
    responses to environmental events.
    
    \item [\textbf{R5}] \textbf{Lightweight edge inference.}
    The model should enable low inference speed and memory-efficient calibration
    on resource-constrained IoT devices.
\end{itemize}

\subsection{Model Design}
\label{subsec:method}
To satisfy R1-R5, we adopt an attention-free MLP architecture. Temporal compression and residual temporal-feature fusion support transferable and shift-robust calibration (R1-R4), while operating in a compact temporal space keeps inference and memory costs low for edge deployment (R5).

\paragraph{Initial Temporal Compression.}

Let \(x_t\) and \(y_t\) denote the measurements of a low-cost sensor and its reference sensor at time \(t\), respectively. 
Given an input window of length \(L\), the calibration input is defined as:
\begin{equation}
\label{eq:calibration_input}
\mathbf{X}_t
=
\left[
\mathbf{s}_{t-L+1},
\ldots,
\mathbf{s}_{t}
\right]^{\top}
\in\mathbb{R}^{L\times F},
\end{equation}
where \(\mathbf{s}_t\in\mathbb{R}^{F}\) has a low-cost sensor
measurement and optional temporal features. 
The model estimates the reference \(y_t\) corresponding to the final time step of the input window.
For notational simplicity, the time index is often omitted, \textit{i.e.}, \(\mathbf{X}=\mathbf{X}_t\). 
The model first projects the \(L\) input time steps into \(R\) compressed temporal representations:
\begin{equation}
\label{eq:temporal_compression}
\mathbf{Z}
=
\left[
\operatorname{ReLU}
\left(
\mathbf{X}^{\top}\mathbf{W}_{p}
+
\mathbf{b}_{p}
\right)
\right]^{\top}\in\mathbb{R}^{R\times F}
\end{equation}
Here,
\(\mathbf{W}_{p}\in\mathbb{R}^{L\times R}\) with \(R\ll L\).
This projection reduces the cost of subsequent operations while retaining the dominant temporal patterns within the input window.
Operating in this compact temporal space reduces sensitivity to short-lived, location-specific fluctuations and promotes more transferable temporal patterns.
This design is intended to improve spatial generalization when the reference location changes (R1) and when the model is transferred to unseen locations (R2). This compression also reduces the cost of subsequent transformations, supporting lightweight edge inference (R5).

\paragraph{Temporal and Feature Fusion.}

The representation \(\mathbf{Z}\) is first transformed along the temporal dimension:
\begin{equation}
\label{eq:temporal_transformation}
\mathbf{\tilde{H}}
=
\operatorname{BN}\!
\left(
\mathbf{Z}
+
\left[
\operatorname{ReLU}\!
\left(
\mathbf{Z}^{\top}\mathbf{W}
+
\mathbf{b}
\right)
\right]^{\top}
\right)
\end{equation}
where
\(\mathbf{W}\in\mathbb{R}^{R\times R}\).
The fusion captures relationships among latent time steps, including gradual trends, delayed sensor responses, and rapid changes.
The model then applies a nonlinear transformation along the feature dimension \(D\):
\begin{equation}
\label{eq:feature_transformation}
\mathbf{H}
\!=\!
\operatorname{BN}\!
\left(
\mathbf{\tilde{H}}
\!+\!
\operatorname{ReLU}\!
\left(
\mathbf{\tilde{H}}\mathbf{W\!}_{1}
\!+\!
\mathbf{b}_{1}
\right)\!
\mathbf{W\!}_{2}
\!+\!
\mathbf{b}_{2}
\right)\!\in\!\mathbb{R}^{R\!\times\!F}
\end{equation}
where
\(\mathbf{W}_{1}\in\mathbb{R}^{F\times D}\) and
\(\mathbf{W}_{2}\in\mathbb{R}^{D\times F}\). 
With multiple input features, this transformation learns nonlinear dependencies among sensor and temporal variables. 
With a single sensor channel, it acts as a lightweight nonlinear calibration.

Residual connections preserve the underlying sensor trajectory while allowing the nonlinear transformations to learn the required correction. 
Batch normalization stabilizes the scale of intermediate representations. 
These components are intended to maintain stable representations under the long-term baseline and response changes considered (R3).

The temporal fusion also processes the recent trajectory rather than relying only on the final measurement. 
When a door or window transition and its resulting sensor response are contained within the input window, the model can represent the direction and duration of the change. 
This property supports calibration under the abrupt events considered (R4).

\paragraph{Calibration Output Head.}

The final representation
\(\mathbf{H}\) is aggregated over the latent temporal and feature dimensions:
\begin{equation}
\label{eq:calibration_output}
\hat{y}
=
\mathbf{w}_{f}^{\top}
\!\left(
\mathbf{H}^{\top}\mathbf{w}_{o}
+
b_{o}\mathbf{1}_{F}
\right)\!
+
b_{f}
\end{equation}
Here, \(\mathbf{w}_{o}\in\mathbb{R}^{R}\) aggregates the latent temporal representation of each feature, while \(\mathbf{w}_{f}\in\mathbb{R}^{F}\) combines the feature-wise values. 
The scalar output \(\hat{y}\) is the calibrated estimate corresponding to the final time step of the input window.

The model does not use location-specific information including particular door or window states.
It therefore learns sensor-response patterns shared across locations instead of explicitly
memorizing individual environments. 
Initial compression
supports R1 and R2, while the residual representation supports long-term drift (R3) and abrupt events (R4).
Moreover, initial compression before the fusion blocks and using small hidden dimensions keep the parameter count and inference cost low (R5). 
This makes it suitable for real-time calibration on resource-constrained edge devices.





\section{Experimental Settings}
\label{sec: 06-Experimental_setting}

This section describes the evaluation protocols, baselines, metrics, and implementation details.

\subsection{Evaluation Objectives}
Our experiments evaluate whether the proposed dataset and four scenarios expose calibration challenges that are obscured by standard strict pairwise evaluation. We further examine whether our model consistently achieves strong calibration performance under limited reference setting, unseen-location transfer, long-term drift, and event-conditioned shifts, while retaining low computational cost. Table~\ref{tb:outdoor_indoor_results} reports standard indoor and outdoor calibration performance and computational efficiency. Tables~\ref{tab:scenario1_sensor_average}-\ref{tab:scenario3_event} report the results for S1-S4, and Figure~\ref{fig:on-device} evaluates latency and model size on the target edge device.

\subsection{Dataset Usage and Evaluation Protocol}
\label{sec:05-experiment-dataset}
Indoor experiments in
Tables~\ref{tab:scenario1_sensor_average}-\ref{tab:scenario3_event}
use the dataset from Section~\ref{sec:dataset}, while the outdoor
experiments in Table~\ref{tb:outdoor_indoor_results} follow
settings~\cite{ahn2025real}. For standard calibration results in
Table~\ref{tb:outdoor_indoor_results}, calibration is performed
separately at each location using
co-located low-cost and reference-sensor measurements, and the
reported root mean squared error (RMSE) is averaged across all locations.

For S1 and S2, samples at each location are chronologically split into 70\% training, 15\% validation, and 15\% test sets. In S1, each location serves as the comparison-sensor-equipped anchor. Training uses low-cost measurements from all locations but reference-channel supervision only from the anchor. For each anchor, RMSE is macro-averaged across the four non-anchor locations and then averaged across all five anchors. In S2, a model is trained on each location and evaluated without adaptation on the other four. RMSE is macro-averaged across the target locations and then across the five source locations.

In S3, each location-sensor time series is chronologically split into 50\% training, 10\% validation, and 40\% deployment evaluation data. The evaluation data are split into four periods, with drift measured by the relative RMSE increase from the first to the last. Results are averaged across the five locations. In S4, window-related periods at the South location and door-related periods at the North location are identified from the number of open windows and the entrance-door state, respectively. Using the same trained model, event-period RMSE is evaluated for CO$_2$ and temperature, which are sensitive to external environmental events.

\begin{table}[t]
\centering
\caption{S1 reports the average RMSE on non-anchor
locations across the five anchor-location settings.}
\label{tab:scenario1_sensor_average}
\vspace{-0.5em}
\resizebox{\columnwidth}{!}{
\setlength\tabcolsep{7pt}
\begin{tabular}{lccccc}
\toprule
\multirow{2.5}{*}{\textbf{Model}}
& \multicolumn{5}{c}{\textbf{RMSE} $\downarrow$} \\
\cmidrule(lr){2-6}
& CO & CO$_2$ & T. & RH. & PM$_1$ \\
\midrule
iTransformer
& 6.26
& \underline{9.45}
& \first{4.11}
& \underline{7.25}
& 9.34 \\

TimeXer
& 7.92 & 10.29 & 9.18 & 9.31 & 12.95 \\

DLinear
& 7.71 & 10.13 & 5.02 & 11.91 & 14.45 \\

XLinear
& 6.93 & 10.77 & 6.59 & 10.80 & 11.69 \\

TimeMixer
& 9.51 & 10.65 & 6.91 & 10.71 & 12.54 \\

SenDaL
& 9.35 & 10.89 & 8.28 & 9.21 & 9.67 \\

TESLA
& \underline{6.19}
& 10.13
& 4.36
& 7.96
& \underline{9.17} \\

\textbf{\textsc{Ours}}
& \first{4.69}
& \first{8.45}
& \underline{4.30}
& \first{6.78}
& \first{8.74} \\
\bottomrule
\end{tabular}
}
\vspace{-0.3em}
\end{table}
\begin{table}[t]
\centering
\caption{S2 evaluates transfer from each source
location to four unseen target locations without adaptation, averaged
across the five source-location settings.}
\vspace{-0.5em}
\label{tab:scenario2_location_transfer}
\resizebox{\columnwidth}{!}{
\setlength\tabcolsep{6.4pt}
\begin{tabular}{lccccc}
\toprule
\multirow{2.5}{*}{\textbf{Model}}
& \multicolumn{5}{c}{\textbf{RMSE} $\downarrow$} \\
\cmidrule(lr){2-6}
& CO
& CO$_2$
& T.
& RH.
& PM$_1$ \\
\midrule
iTransformer
& 7.63
& 10.45
& \underline{4.79}
& 9.02
& 10.27 \\

TimeXer
& 9.35
& 11.18
& 6.66
& 10.34
& 11.88 \\

DLinear
& 8.69
& 11.60
& 11.43
& 12.27
& 12.66 \\

XLinear
& 8.44
& 11.81
& 7.62
& 10.06
& 12.56 \\

TimeMixer
& 9.70
& 11.75
& 8.54
& 13.05
& 13.65 \\

SenDaL
& 9.48
& 10.92
& 5.94
& 10.34
& 12.04 \\

TESLA
& \first{7.44}
& \underline{10.43}
& 5.15
& \underline{8.58}
& \underline{9.17} \\

\textbf{\textsc{Ours}}
& \underline{7.56}
& \first{9.68}
& \first{4.51}
& \first{8.21}
& \first{8.85} \\
\bottomrule
\end{tabular}
}
\vspace{-0.3em}
\end{table}
\begin{table}[t]
\centering
\caption{S3 average relative increase from the
earliest to the latest deployment period across five locations.}
\vspace{-0.5em}
\label{tab:scenario4_drift}
\resizebox{\columnwidth}{!}{
\setlength\tabcolsep{7.5pt}
\begin{tabular}{lccccc}
\toprule
\multirow{2.5}{*}{\textbf{Model}}
& \multicolumn{5}{c}{
    \textbf{Relative increase} (\%) $\downarrow$
} \\
\cmidrule(lr){2-6}
& CO
& CO$_2$
& T.
& RH.
& PM$_1$ \\
\midrule
iTransformer
& \underline{28.6}
& 56.0
& \underline{32.3}
& 47.6
& 67.8 \\

TimeXer
& 64.4
& \underline{42.2}
& 53.7
& 67.2
& 84.5 \\

DLinear
& 63.8
& 84.9
& 49.2
& 65.1
& 108.8 \\

XLinear
& 43.3
& 82.4
& 56.8
& 75.2
& 77.6 \\

TimeMixer
& 64.2
& 73.5
& 61.1
& 60.1
& 97.5 \\

SenDaL
& 87.1
& 78.6
& 62.4
& 105.9
& 75.2 \\

TESLA
& 43.4
& 56.8
& 59.2
& \underline{46.5}
& \underline{64.6} \\

\textbf{\textsc{Ours}}
& \first{24.6}
& \first{37.0}
& \first{32.2}
& \first{30.6}
& \first{57.2} \\
\bottomrule
\end{tabular}
}
\vspace{-0.2em}
\end{table}
\begin{table}[t]
\centering
\caption{S4 event-conditioned calibration results at
the window/door-side locations. Each value reports RMSE
during the corresponding event periods.}
\vspace{-0.5em}
\label{tab:scenario3_event}
\resizebox{\columnwidth}{!}{
\setlength\tabcolsep{5pt}
\begin{tabular}{lccccc}
\toprule
\multirow{2.5}{*}{\textbf{Model}}
& \multicolumn{2}{c}{\textbf{South (Window)}}
& \multicolumn{2}{c}{\textbf{North (Door)}}
& \multirow{2.5}{*}{\textbf{Avg.}} \\
\cmidrule(lr){2-3}
\cmidrule(lr){4-5}
& CO$_2$ & T.
& CO$_2$ & T.
& \\
\midrule
iTransformer
& 18.48
& 7.97
& 19.26
& 7.29
& 13.25 \\

TimeXer
& 21.69
& 9.49
& 24.33
& 10.68
& 16.55 \\

DLinear
& 20.55
& 10.37
& 24.08
& 10.93
& 16.48 \\

XLinear
& 23.74
& 8.05
& 26.55
& 10.48
& 17.20 \\

TimeMixer
& 19.21
& 10.58
& 32.08
& 11.69
& 18.39 \\

SenDaL
& 20.68
& 10.61
& \first{15.35}
& 9.79
& 14.11 \\

TESLA
& \underline{17.70}
& \first{5.54}
& 21.07
& \underline{6.27}
& \underline{12.64} \\

\textbf{\textsc{Ours}}
& \first{17.60}
& \underline{5.55}
& \underline{15.51}
& \first{5.78}
& \first{11.11} \\
\bottomrule
\end{tabular}
}
\vspace{-0.5em}
\end{table}

\subsection{Baselines}

We compare the proposed method with representative attention-based,
linear/MLP-based, and calibration-specific models. The baselines include
iTransformer \cite{liu2024itransformer} and TimeXer
\cite{wang2024timexer}, followed by DLinear
\cite{zeng2023dlinear}, XLinear \cite{chen2026xlinear}, and
TimeMixer \cite{wang2023timemixer}. We further include SenDaL
\cite{ahn24sendal} and TESLA \cite{ahn2025real}, which were
specifically developed for efficient sensor calibration. Veli
\cite{dalbah2026veli} is excluded because it is unsupervised, unlike
our supervised benchmark.

\subsection{Evaluation Metrics}

Calibration accuracy is measured using RMSE in the original physical unit of each sensor variable. RMSE penalizes
large deviations, which are particularly relevant under long-term drift
and abrupt environmental changes. Computational efficiency is evaluated using per-sample inference time
and peak inference memory. Inference time is averaged over repeated
runs after graph warm-up using a fixed benchmark batch. Peak memory is
measured relative to the pre-inference process baseline.

\subsection{Implementation Details}

All models are implemented in TensorFlow 2.14 and trained under a
common protocol adopted from \citet{ahn2025real}, following the
iTransformer configuration \cite{liu2024itransformer}. Efficiency is measured under identical conditions on an
AMD EPYC 7513 32-Core Processor. For embedded evaluation, the trained
models are converted to TensorFlow Lite FlatBuffers and deployed on an
Arduino Nano 33 BLE Sense.

\begin{figure}[t]
    \centering
    \vspace{-0.5em}
    \includegraphics[width=0.99\columnwidth]{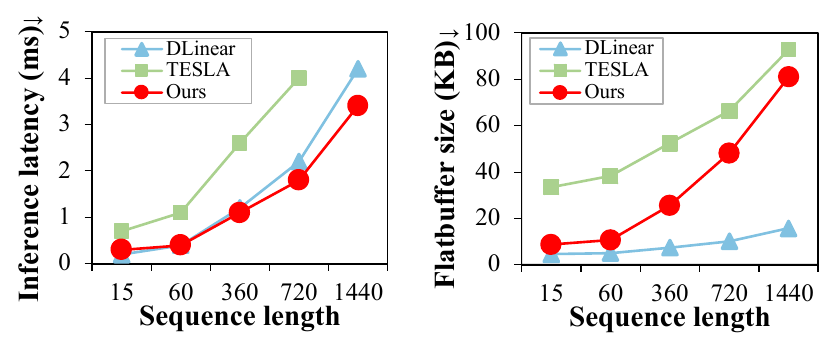}
    \vspace{-0.5em}
    \caption{Performance on Arduino Nano 33 BLE Sense. The graphs show
  mean inference latency and FlatBuffer size across input windows of
  15, 60, 360, 720, and 1440. Missing points indicate that on-device
  measurement was not possible.}
    \label{fig:on-device}
\end{figure}

\section{Experimental Results}
\label{sec_result}

This section presents calibration and efficiency results under standard and four deployment-oriented scenario.

\subsection{Overall Calibration Performance}

Table~\ref{tb:outdoor_indoor_results} compares calibration performance
on SensEURCity and the proposed indoor dataset. The proposed model
achieved the lowest RMSE for all three outdoor variables and four of
the five indoor variables, while ranking second for indoor temperature.
These results establish strong performance under co-located reference
supervision.

\subsection{R1. Reference-Efficient Calibration}

Table~\ref{tab:scenario1_sensor_average} shows that the proposed model
achieved the lowest RMSE for CO, CO$_2$, humidity, and PM$_1$.
TESLA outperformed iTransformer on four indoor variables under strict
pairwise calibration, whereas iTransformer performed better for
CO$_2$, temperature, and humidity in S1. This rank reversal shows that overall calibration performance does not necessarily imply
effective sharing of limited reference supervision.

\subsection{R2. Location-Transfer Calibration}

Table~\ref{tab:scenario2_location_transfer} shows higher RMSE than S1
for all five variables of the proposed model, reflecting the difficulty
of transfer to unseen locations. Nevertheless, the model achieved the
best results for CO$_2$, temperature, humidity, and PM$_1$, and ranked
second for CO. The S1-to-S2 increase varied from 1.3\% for PM$_1$ to
61.2\% for CO, showing that reference-efficient learning and
unseen-location transfer impose distinct challenges.

\subsection{R3. Long-Term Calibration Drift}

Table~\ref{tab:scenario4_drift} shows positive RMSE increases for all
models and variables, demonstrating persistent temporal degradation over
the six-month deployment. The proposed model achieved the lowest
relative increase for all five variables, although its RMSE still
rise by 24.6-57.2\%. Despite outperforming iTransformer
under standard calibration, TESLA showed larger degradation for CO,
CO$_2$, and temperature. Thus, standard accuracy does not necessarily
indicate long-term robustness.

\subsection{R4. Event-Conditioned Calibration}

Table~\ref{tab:scenario3_event} shows that the best-performing model
varied across event conditions. TESLA achieved the lowest RMSE for
South-window temperature, whereas SenDaL performed best for North-door
CO$_2$. The proposed model performed best for the other two conditions
and ranked first or second in all four cases. These condition-specific
rankings show that aggregate results can obscure location- and
modality-dependent event responses.

\subsection{R5. Efficiency and On-device Performance}

Figure~\ref{fig:on-device} shows that the proposed model
achieved the fastest inference time with low peak memory. We compare
DLinear, the fastest on-device linear baseline, and TESLA, a recent
calibration transformer model~\cite{ahn2025real}. On the Arduino
Nano 33 BLE Sense, it was faster than TESLA at all sequence lengths and
than DLinear for long input windows. Its FlatBuffer size was larger than
DLinear but smaller than TESLA, which failed at the longest sequence
length. These results demonstrate efficient long-window calibration on
resource-constrained devices.

\section{Discussion and Limitations}
\label{sec:discussion}
We discuss the implications of our work.
\subsection{Dataset and Scenario}

Model rankings changed with reference coverage, target-location
exposure, deployment time, and event conditions. This shows that
location-specific accuracy, reference-efficient learning,
unseen-location transfer, long-term robustness, and event robustness
are distinct calibration capabilities. The proposed scenarios therefore
reveal deployment challenges hidden by standard aggregate evaluation.
The synchronized multi-location measurements, six-month time series,
and event metadata support complementary evaluation of spatial and
temporal robustness.

\subsection{Model and Deployment}

Our model ranked first or second across the deployment scenarios while
maintaining low edge-inference cost. These results indicate robust
calibration across locations, time periods, and event conditions without
sacrificing efficiency. However, the 24.6-57.2\% degradation in S3
shows that static calibration cannot eliminate long-term drift and
motivates periodic recalibration or online adaptation. The on-device
results further demonstrate efficient inference with input windows of
up to 1,440 time steps.

\subsection{Single-Environment Scope and Data Fidelity}

The indoor dataset was collected at five locations within one building,
limiting its coverage of other buildings, climates, and ventilation
conditions. However, this controlled setting enabled dense,
synchronized, and long-term measurements under a consistent hardware
and sampling protocol. It also supported uniform annotation of door
state, window status, occupancy estimates, and business hours. Because
the metadata were recorded at fixed intervals rather than exact event
times, S4 evaluates event-conditioned periods rather than precisely
aligned event responses.

\section{Conclusion}
\label{sec:07-conclusion}

We presented a deployment-oriented benchmark for indoor low-cost sensor calibration. We constructed a six-month, five-location dataset and defined four scenarios covering limited reference availability, unseen locations, long-term drift, and event-conditioned shifts. Based on these scenarios, we derived design requirements and proposed a lightweight temporal model for robust and efficient calibration. Experimental results demonstrate strong performance across the four scenarios with low edge-inference cost. Future work will extend the dataset to diverse indoor environments and explore online adaptation for long-term drift.


\appendix


\bibliography{Bibliography-File}


\end{document}